\documentclass{llncs}
\usepackage{amssymb}
\usepackage{amsmath}
\usepackage{graphicx}
\usepackage{hyperref}
\usepackage{url}

\usepackage{float}
\floatstyle{plaintop}
\restylefloat{table}

\newcommand{\keywords}[1]{\par\addvspace\baselineskip
\noindent\keywordname\enspace\ignorespaces#1}

\usepackage{array}
\newcolumntype{C}[1]{>{\centering\let\newline\\\arraybackslash\hspace{0pt}}m{#1}}
\newcolumntype{L}[1]{>{\let\newline\\\arraybackslash\hspace{0pt}}m{#1}}

\begin{document}

\mainmatter  

\title{Plant identification based on noisy web data: the amazing performance of deep learning (LifeCLEF 2017)}

\titlerunning{LifeCLEF Plant Identification Task 2017}

\author{Herv\'e Go\"eau\inst{1}
  \and Pierre Bonnet\inst{1}
  \and Alexis Joly\inst{2,3}
}

\tocauthor{Herv\'e Go\"eau, Alexis Joly, Pierre Bonnet}

\institute{CIRAD, UMR AMAP, France,
\email{herve.goeau@cirad.fr, pierre.bonnet@cirad.fr}
\and 
Inria ZENITH team, France, 
\email{alexis.joly@inria.fr}
\and
LIRMM, Montpellier, France
}

\toctitle{LifeCLEF Plant Identification Task 2017}

\maketitle

\begin{abstract}
The 2017-th edition of the LifeCLEF plant identification challenge is an important milestone towards automated plant identification systems working at the scale of continental floras with 10.000 plant species living mainly in Europe and North America illustrated by a total of 1.1M images. Nowadays, such ambitious systems are enabled thanks to the conjunction of the dazzling recent progress in image classification with deep learning and several outstanding international initiatives, such as the Encyclopedia of Life (EOL), aggregating the visual knowledge on plant species coming from the main national botany institutes. However, despite all these efforts the majority of the plant species still remain without pictures or are poorly illustrated. Outside the institutional channels, a much larger number of plant pictures are available and spread on the web through botanist blogs, plant lovers web-pages, image hosting websites and on-line plant retailers. The LifeCLEF 2017 plant challenge presented in this paper aimed at evaluating to what extent a large noisy training dataset collected through the web and containing a lot of labelling errors can compete with a smaller but trusted training dataset checked by experts. To fairly compare both training strategies, the test dataset was created from a third data source, \textit{i.e.} the Pl@ntNet mobile application that collects millions of plant image queries all over the world. This paper presents more precisely the resources and assessments of the challenge, summarizes the approaches and systems employed by the participating research groups, and provides an analysis of the main outcomes.
\end{abstract}

\section{Introduction} 
 Thanks to the long term efforts made by the biodiversity informatics community, it is now possible to aggregate validated data about tens of thousands species world wide. The international initiative Encyclopedia of Life (EoL), in particular, is one of the largest repository of plant pictures. However, despite these efforts, the majority of plant species living on earth are still very poorly illustrated with typically only few pictures of a single specimen or herbarium scans. More pictures are available on the Web through botanist blogs, plant lovers web-pages, image hosting websites and on-line plant retailers. But this data is much harder to be structured and contains a high degree of noise. The LifeCLEF 2017 plant identification challenge proposes to study to what extent such noisy web data is competitive with a relatively smaller but trusted training set checked by experts. As a motivation, a previous study conducted by Krause et al. \cite{krause2016unreasonable} concluded that training deep neural networks on noisy data was unreasonably effective for fine-grained recognition. The PlantCLEF challenge completes their work in several points:
\begin{enumerate}
    \item it extends their result to the plant domain whose specificity is that the available data on the web is scarcer, the risk of confusion higher and, finally, the degree of noise higher.
    \item it scales the comparison between trusted and noisy training data to 10K of species whereas the trusted training sets used in their study were actually limited to few hundreds of species. 
    \item it uses a third-party test dataset that is not a subset of either the noisy dataset or the trusted dataset. This allows a more fair comparison. More precisely, the test data is composed of images submitted by the crowd of users of the mobile application Pl@ntNet\cite{joly2016look}. Consequently, it exhibits different properties in terms of species distribution, pictures quality, etc.
\end{enumerate}

In the following subsections, we synthesize the resources and assessments of the challenge, summarize the approaches and systems employed by the participating research groups, and provide an analysis of the main outcomes.

\section{Dataset}

To evaluate the above mentioned scenario at a large scale and in realistic conditions, we built and shared three datasets coming from different sources. As training data, in addition to the data of the previous PlantCLEF challenge \cite{goeau2016plant}, we provided two new large data sets both based on the same list of 10,000 plant species (living mainly in Europe and North America):\\
\\
\textbf{Trusted Training Set \textit{EoL10K}}: a trusted training set based on the online collaborative Encyclopedia Of Life (EoL). The 10K species were selected as the most populated species in EoL data after a curation pipeline (taxonomic alignment, duplicates removal, herbaria sheets removal, etc.). The training set contains 256,287 pictures in total but has a strong class imbalance with a minimum of 1 picture for \textit{Achillea filipendulina} and a maximum of 1245 pictures for \textit{Taraxacum laeticolor}.\\
\\
\textbf{Noisy Training Set \textit{Web10K}}: a noisy training set built through Web crawlers (Google and Bing image search engines) and containing 1.1M images. This training set is also imbalanced with a minimum of 4 pictures for \textit{Plectranthus sanguineus} and a maximum of 1732 pictures for \textit{Fagus grandifolia}. 
\\
The main objective of providing both datasets was to evaluate to what extent machine learning techniques can learn from noisy data compared to trusted data (as usually done in supervised classification). Pictures of EoL are themselves coming from different sources, including institutional databases as well as public data sources such as Wikimedia, iNaturalist, Flickr or various websites dedicated to botany. This aggregated data is continuously revised and rated by the EoL community so that the quality of the species labels is globally very good. On the other side, the noisy web dataset contains more images but with several types and levels of noise: some images are labeled with the wrong species name (but sometimes with the correct genus or family), some are portraits of a botanist specialist of the targeted species, some are labeled with the correct species name but are drawings or herbarium sheets, etc.\\
\\
\textbf{Pl@ntNet test set}: the test data to be analyzed within the challenge is a large sample of the query images submitted by the users of the mobile application Pl@ntNet (iPhone\footnote{\url{https://itunes.apple.com/fr/app/plantnet/id600547573?mt=8}} \& Androïd\footnote{\url{https://play.google.com/store/apps/details?id=org.plantnet}}). It contains a large number of wild plant species mostly coming from the Western Europe Flora and the North American Flora, but also plant species used all around the world as cultivated or ornamental plants including some endangered species.\\
\\

\section{Task Description}
Based on the previously described testbed, we conducted a system-oriented evaluation involving different research groups who downloaded the data and ran their system. 

Each participating group was allowed to submit up to 4 \textit{run files} built from different methods (a \textit{run file} is a formatted text file containing the species predictions for all test items). Semi-supervised, interactive or crowdsourced approaches were allowed but had to be clearly signaled within the submission system. None of the participants employed such methods.\\
\\
The main evaluation metric is the Mean Reciprocal Rank (MRR), a statistic measure for evaluating any process that produces a list of possible responses to a sample of queries ordered by probability of correctness. The reciprocal rank of a query response is the multiplicative inverse of the rank of the first correct answer. The MRR is the average of the reciprocal ranks for the whole test set:
$$MRR = \frac{1}{|Q|} \sum_{i=1}^{Q}\frac{1}{rank_i}$$
where $|Q|$ is the total number of query occurrences in the test set.

\section{Participants and methods}
80 research groups registered to the LifeCLEF plant challenge 2017. Among this large raw audience, 8 research groups finally succeeded in submitting run files. Details of the used methods and evaluated systems are synthesized below and further developed in the working notes of the participants (CMP \cite{CMP2017}, FHDO BCSG \cite{FHDOBCSG2017}, KDE TUT \cite{KDETUT2017}, Mario MNB \cite{Mario2017}, Sabanci Gebze\cite{Sabanci2017}, UM \cite{UM2017} and UPB HES SO \cite{UPBHESSO2017}). Table \ref{tab:rawresults} reports the results achieved by each run as well as a brief synthesis of the methods used in each of them. Complementary, the following paragraphs give a few more details about the methods and the overall strategy employed by each participant.\\
\\
\begin{table}
    \centering
    \vspace{3mm}
    \begin{tabular}{|C{25mm}|C{36mm}|C{32mm}|C{9mm}|C{9mm}|C{9mm}|}
    \hline
    Run & Method & Training & MRR & Top1 & Top5\\
    \hline
    \hline
Mario TSA Berlin Run4
& Average of many fine-tuned NNs
& EOL, WEB
& 0.92 & 0.885 & 0.962\\
\hline
Mario TSA Berlin Run2
& Average of 6 fine-tuned NNs
& EOL, WEB, PlantCLEF2016
& 0.915	& 0.877	& 0.96\\
\hline
Mario TSA Berlin Run3
& Average of 3 fine-tuned NNs
& EOL, filtered WEB, PlantCLEF2016, high scoring test data
& 0.894	& 0.857 & 0.94\\
\hline
KDE TUT Run4 
& ResNet50 (modified)
& EOL, WEB
& 0.853 & 0.793 & 0.927\\
\hline
Mario TSA Berlin Run3
& Average of 3 fine-tuned NNs
& EOL, PlantCLEF2016
& 0.847 & 0.794	& 0.911\\
\hline
CMP Run1
& Inception-ResNet-v2 
& EOL, filtered WEB
& 0.843	& 0.786 &	0.913\\
\hline
KDE TUT Run3 
& ResNet50 (modified)
& EOL, WEB
& 0.837	& 0.769 & 0.922\\
\hline
CMP Run3 
& Inception-ResNet-v2
& EOL
& 0.807 & 0.741	& 0.887\\
\hline
FHDO BCSG Run2
& Inception-ResNet-v2
& EOL, filtered WEB
& 0.806	& 0.738	& 0.893\\
\hline
FHDO BCSG Run3
& Inception-ResNet-v2
& EOL, filtered WEB
& 0.804	& 0.736 & 0.891\\
\hline
UM Run2 
& VGGNet
& WEB
& 0.799 & 0.726 & 0.888\\
\hline
UM Run3 
& VGGNet multi-organ 
& EOL, WEB
& 0.798	& 0.727 & 0.886\\
\hline
FHDO BCSG Run1
& Inception-ResNet-v2
& EOL 
& 0.792 & 0.723 & 0.878\\
\hline
UM Run4 
& UM Run1\&2 max voting
& EOL, WEB
& 0.789 & 0.715 & 0.882\\
\hline
KDE TUT Run1
& ResNet50 (modified)
& EOL
& 0.722	& 0.707	& 0.85\\
\hline
CMP Run2 
& Inception-ResNet-v2
& EOL, filtered WEB
& 0.765	& 0.68 & 0.87\\
\hline
CMP Run4 
& Inception-ResNet-v2
& EOL
& 0.733 & 0.641	& 0.849\\
\hline
UM Run1
& VGGNet multi-organ
& EOL
& 0.7 & 0.621	& 0.795\\
\hline
SabanciU GebzeTU Run 4
& VGGNets
& EOL, filtered WEB
& 0.638 & 0.557 & 0.738\\
\hline
SabanciU GebzeTU Run 1
& VGGNets 
& EOL, filtered WEB
& 0.636 & 0.556 & 0.737\\
\hline
SabanciU GebzeTU Run 3
& VGGNets 
& EOL, filtered WEB
& 0.622	& 0.537 & 0.728\\
\hline
PlantNet Run1
& Inception v1
& EOL
& 0.613 & 0.513	& 0.734\\
\hline
SabanciU GebzeTU Run2
& VGGNets
& EOL
& 0.581 & 0.508	& 0.68\\
\hline
UPB HES SO Run3
& AlexNet
& EOL
& 0.361 & 0.293	& 0.442\\
\hline
UPB HES SO Run4
& AlexNet
& EOL
& 0.361 & 0.293	& 0.442\\
\hline
UPB HES SO Run1
& AlexNet
& EOL
& 0.326 & 0.26 & 0.406\\
\hline
UPB HES SO Run2
& AlexNet
& EOL
& 0.305 & 0.239	& 0.383\\
\hline
FHDO BCSG Run4
& Inception v4
& PlantCLEF2016, WEB
& 0 & 0	& 0\\
\hline
\end{tabular}
\caption{Results of the LifeCLEF 2017 Plant Identification Task}
\label{tab:rawresults}
\end{table}
\textbf{CMP, Czech Republic, 4 runs, \cite{CMP2017}}: this participant based his work on the Inception-ResNet-v2 architecture \cite{szegedy2016inception} which introduces inception modules with residual connections. An additional maxout fully-connected layer with batch normalization was added on top of the network, before the classification fully-connected layer. Hard bootstrapping was used for training with noisy labels. A total of 17 models were trained using different training strategies such as: with or without maxout, with or without pre-training on ImageNet, with or without bootstrapping, with and without filtering of the noisy web dataset. CMP Run 1 is the combination of all the 17 networks by averaging their results. CMP Run 3 is the combination of the 8 networks that were trained on the trusted EOL data solely. CMP Run2 and CMP Run 4 are post-processings of CMP Run1 and CMP Run 3 aimed at compensating the asymmetry of class distributions between the test set and the training sets.\\
\\
\textbf{FHDO BCSG, Germany, 4 runs, \cite{FHDOBCSG2017}}: this participant also used Inception-ResNet-v2 architecture. The Run 1 is based exclusively on the trusted EOL dataset following a two phases fine-tuning approach: in a first phase only the last output layer is trained for a few epochs with a small learning rate starting from randomly initialized weights, and in a second phase the entire network is trained with numerous epochs and a larger learning rate. For the test set, they used an oversampling technique for increasing the number of test samples with 10 crops (1 center, 4 corners and the mirrored crops). For the Run 2, they kept the same architecture but extended the trusted dataset with a filtered subset of the noisy dataset: images from the web were added if their species label were in the top-5 predictions from the model used in Run 1. Run 3 is the combination of Run 1\&2.\\
\\
\textbf{KDE TUT, Japan, 4 runs, \cite{KDETUT2017}}: this participant introduced a modified version of the ResNet-50 model. Three of the intermediate convolutional layers used for downsampling were modified by changing the stride value from 2 to 1 and preceding it by max-pooling with a stride of 2, to optimize the coverage of the inputs. Additionally, they switched the downsampling operation with the convolution for delaying the downsampling operation. This has been shown to improve performance by the authors of the ResNet architecture themselves. During the training they used data augmentation based on random crops, rotations and optional horizontal flipping. Test images were also augmented through a single flip operation and the resulting predictions averaged. Since the original ResNet-50 architecture was modified, no fine-tuning was used and the weights were learned from scratch starting with a big learning rate value of 0.1. The learning rates were dropped twice (to 0.01 and then 0.01) over 100 epochs according to a schedule ratio 4:2:1 indicating the number of iterations using the same learning rate (limited to a total number of 350 000 iterations in the case of the big noisy dataset due to technical limitations). Run 1, 2, 3 were trained respectively on the trusted dataset, noisy dataset, and both datasets. The final run 4 is a combination the outputs of the 3 runs.\\
\\
\textbf{PlantNet, France, 1 run}: The PlantNet team provided a baseline for the task with the system used in Pl@ntNet app, based on a slightly modified version of inception v1 model \cite{szegedy2015going} as described in \cite{affouard2017pl}. The system also includes a number of thresholding and rejection mechanisms that are useful within the mobile app but that also degrade the raw classification performance. This team submitted only one run trained on the trusted EOL dataset.\\
\\
\textbf{Sabanci-Gebze, Turkey, 4 runs, \cite{Sabanci2017}}: inspired by the good results achieved last year with a combination of a GoogleLeNet \cite{szegedy2015going} and a VGGNet \cite{Simonyan14c}, this team ran an ensemble classifier of 9 VGGNets. Each network was trained with data augmentation techniques using random crops and random horizontal, focusing only the two last layers for fine-tuning due to technical limitations. The submitted Run 2 used models learned only on the EOL trusted dataset. For the remaining runs, the models were trained for supplementary epochs introducing complementary training images selected from the noisy dataset (about 60.000 images which were matching the ground truth according to the models trained for the Run 2). Run 1, 3, 4 used respectively a Borda count, a maximum confident rule and a weighted combination of the output of the classifiers.\\
\\
\textbf{Mario TSA Berlin, Germany, 4 runs, \cite{Mario2017}}: this participant used ensembles of fine-tuned CNNs pre-trained on ImageNet based on 3 architectures (GoogLeNet, ResNet-152 and ResNeXT) each trained with bagging techniques. Data augmentation techniques multiplied by 5 the number of training images with random cropping, horizontal flipping, variations of saturation, lightness and rotation, these three last transformations following a decreasing degree correlated the diminution of the learning rate during training to let the CNNs see patches closer to the original image at the end of each training process. Test images were also augmented and the resulting predictions averaged. \textit{MarioTsaBerlin Run 1} results from the combination of the 3 architectures trained on the trusted datasets only (EOL and PlantCLEF2016). Run 2 exploited both the trusted and the noisy dataset to train four GoogLeNet's, one ResNet-152 and one ResNeXT. In Run3, two additional GoogLeNet's and one ResNeXT were trained using a filtered version of the web dataset and images of the test set that received a probability higher than 0.98 in Run1. The last and "winning" run \textit{MarioTsaBerlin Run 4} finally combined all the 12 trained models.\\
\\
\textbf{UM, Malaysia \& UK, 4 runs, \cite{UM2017}}: this participant proposed an original architecture called Hybrid Generic-Organ (HBO-CNN) that was trained on the trusted dataset (UM Run 1). Unfortunately, it performed worst than a standard VGGNet model learned on the noisy dataset (UM Run 2). This can be partially explained by the fact that the HBO-CNN model need tagged images (flower, fruit, leaf,...), a missing information for the noisy dataset and partially available for the trusted dataset.\\
\\
\textbf{UPB HES SO, Switzerland, 4 runs, \cite{UPBHESSO2017}}: this team trained the historical AlexNet model \cite{krizhevsky2012imagenet} by using exclusively the trusted training dataset, and focused the experiments on the solver part. Run 1 didn't used weight decay for the regularization. Run 2 applied an important learning rate factor on the last layer without updating this value during the training. Run 3 and 4 used a usual learning rate schedule.

\section{Results}
\begin{figure}[!t]
\centering
\includegraphics[width=0.95\linewidth]{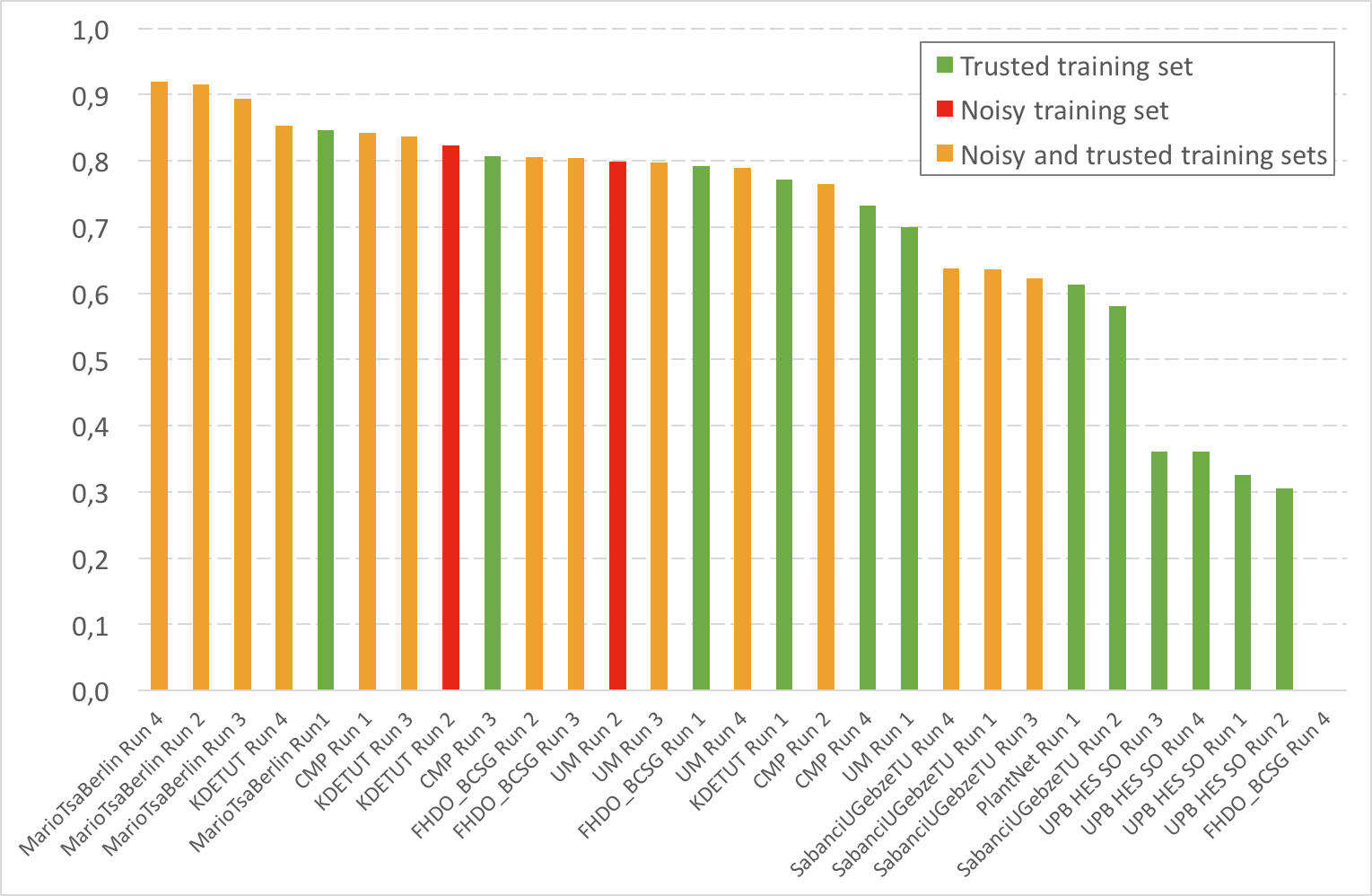}
\caption{Scores achieved by all systems evaluated within the plant identification task of LifeCLEF 2017}
\label{fig:PlantCLEF2017OfficialScore}
\end{figure}

We report in Figure \ref{fig:PlantCLEF2017OfficialScore} the performance achieved by the 29 collected runs. Table \ref{tab:rawresults} provides the results achieved by each run as well as a brief synthesis of the methods used in each of them. 
\\
\\
\textbf{Trusted or noisy ?} As a first noticeable remark, the measured performances are very high despite the difficulty of the task with a median Mean Reciprocal Rank (MRR) around 0.8, and a highest MRR of 0.92 for the best system Mario MNB Run 4. A second important remark is that the best results were obtained mostly by systems that were learned on both the trusted and the noisy datasets. Only two runs (KDE TUT Run 2 and UM Run 2) used exclusively the noisy dataset but gave better results than most of the methods using only the trusted dataset. Several teams also tried to filter the noisy dataset, based on the prediction of a preliminary system trained only on the trusted dataset (\textit{i.e.} by rejecting pictures whose label is contradictory with the prediction). However, this strategy did not improve the final predictor and even degraded the results. For instance Mario MNB Run 2 (using the raw Web dataset) performed better than Mario MNB Run 3 (using the filtered Web dataset).\\
\\
\textbf{Succeeding strategies with CNN models}: Regarding the used methods, all submitted runs were based on Convolutional Neural Networks (CNN) confirming definitively the supremacy of this kind of approach over previous methods. A wide variety of popular architectures were trained from scratch or fine-tuned from pre-trained weights on the ImageNet dataset: GoogLeNet\cite{szegedy2015going} and its improved inception v2\cite{DBLP:journals/corr/IoffeS15} and v4 \cite{szegedy2016inception} versions, inception-resnet-v2\cite{szegedy2016inception}, ResNet-50 and ResNet-152 \cite{he2016deep}, ResNeXT\cite{he2016deep}, VGGNet\cite{Simonyan14c} and even the AlexNet\cite{krizhevsky2012imagenet}. One can notice that inception v3 was not experimented despite the fact that it is a recent model giving state of art performances in other image classification benchmarks. It is important to notice that the best results of each team were obtained with classifier ensembles (in particular Mario TSA Run 4, KDE TUT Run 4 and CMP Run 1). Bootstrap aggregating (bagging) was very efficient in this context to extend the number of classifiers by learning several models with the same architecture but on different training and validation subsets. This is the case of the best run Mario TSA Run 4 that combined 7 GoogLeNet, 2 ResNet-152, 3 ResNeXT trained on different datasets. The CMP team also combined numerous models (up to 17 in Run 1) with various subsets of the training data and bagging strategies, but all with the same inception-resnet-v2 architecture.
Another key for succeeding the task was the use of data augmentation with usual transformations such as random cropping, horizontal flipping, rotation, for increasing artificially the number of training samples and helping the CNNs to generalize better. The two best teams used data augmentation in both the training and the test phase. Mario MNB team added two more interesting transformations, \textit{i.e.} color saturation and lightness modifications. They also correlated the intensity of these transformations with the diminution of the learning rate during training to let the CNNs see patches closer to the original image at the end of each training process. Last but not least, Mario MNB is the only team who used exactly the same transformation in the training and test phase.
Besides the use of ensemble of classifiers, some teams also tried to propose modifications of existing models. KDE TUT, in particular, modified the architecture of the first convolutional layers of ResNet-50 and report consistent improvements in their validation experiments \cite{KDETUT2017}. CMP also reported slight improvements on the inception-resnet-v2 by using a maxout activation function instead of RELU. The UM team proposed an original architecture called Hybrid Generic-Organ learned on the trusted dataset (UM Run 1). Unfortunately, it performed worst than a standard VGGNet model learned on the noisy dataset (UM Run 2). This can be partially explained by the fact that the HBO-CNN model need tagged images (flower, fruit, leaf,...), a missing information for the noisy dataset and partially available for the trusted dataset.\\ 
\\
\textbf{The race for the most recent model?}: one can suppose that the most recent models such as inception-resnet-v2 or inception-v4 should lead to better results than older ones such as AlexNet, VGGNet and GoogleNet. For instance, the runs with GoogleNet and VGGNet by Sabanci \cite{Sabanci2017}, or with a PReLU version of inception-v1 by the PlantNet team, or with the historical AlexNet architecture by the UPB HES SO team \cite{UPBHESSO2017} performed the worst results. However, one can notice that the "winning" team used also numerous GoogLeNet models, while the old VGGNet used in UM run 2 gave quite high and intermediate results around a MRR of 0.8. This highlights how much the training strategies are important and how ensemble classifiers, bagging and data augmentation can greatly improve the performance even without the most recent architectures from the state of the art.\\
\\
\textbf{The race for GPUs}: Like discussed above, best performances were obtained with ensembles of very deep networks trained over millions of images and heavy data augmentation techniques. In the case of the best run Mario MNB Run 4, test images were also augmented so that the prediction of a single image finally relies on the combination of 60 probability distributions (5 patches x 12 models). Overall, the best performing system requires a huge GPU consumption so that their use in data intensive contexts is limited by cost issues (\textit{e.g.} the Pl@ntNet mobile application accounts for millions of users). A promising solution towards this issue could be to rely on knowledge distilling \cite{hinton2015distilling}. Knowledge distilling consists in transferring the generalization ability of a cumbersome model to a small model by using the class probabilities produced by the cumbersome model as soft targets for training the small model. Alternatively, more efficient architectures and learning procedures should be devised.

\section{Complementary Analysis}
\begin{figure}[!t]
\centering
\includegraphics[width=0.95\linewidth]{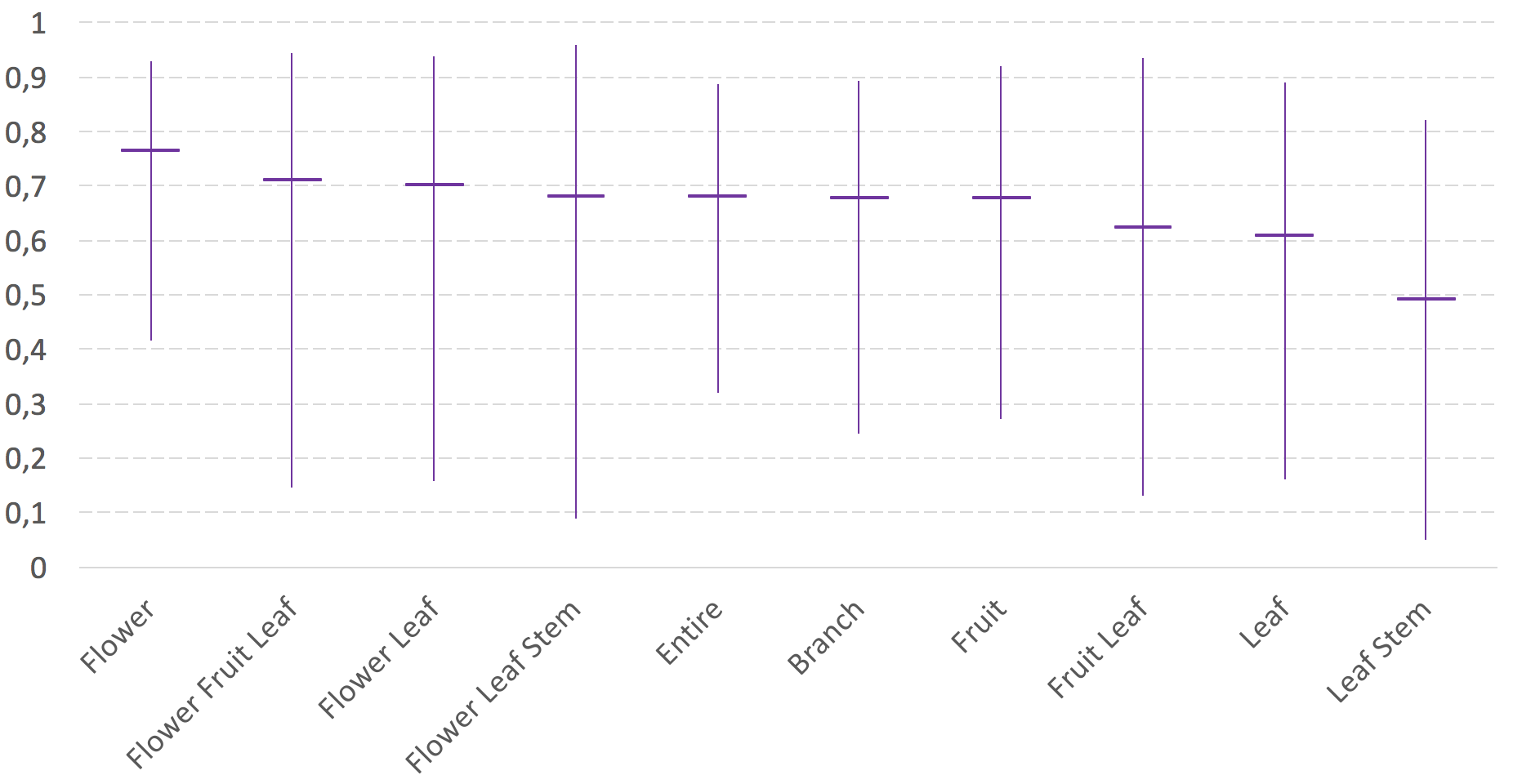}
\caption{Min, max and average MRR scores detailed by organ combination}
\label{fig:PlantCLEF2017ScoresOrgans}
\end{figure}

\begin{figure}[!t]
\centering
\includegraphics[width=0.95\linewidth]{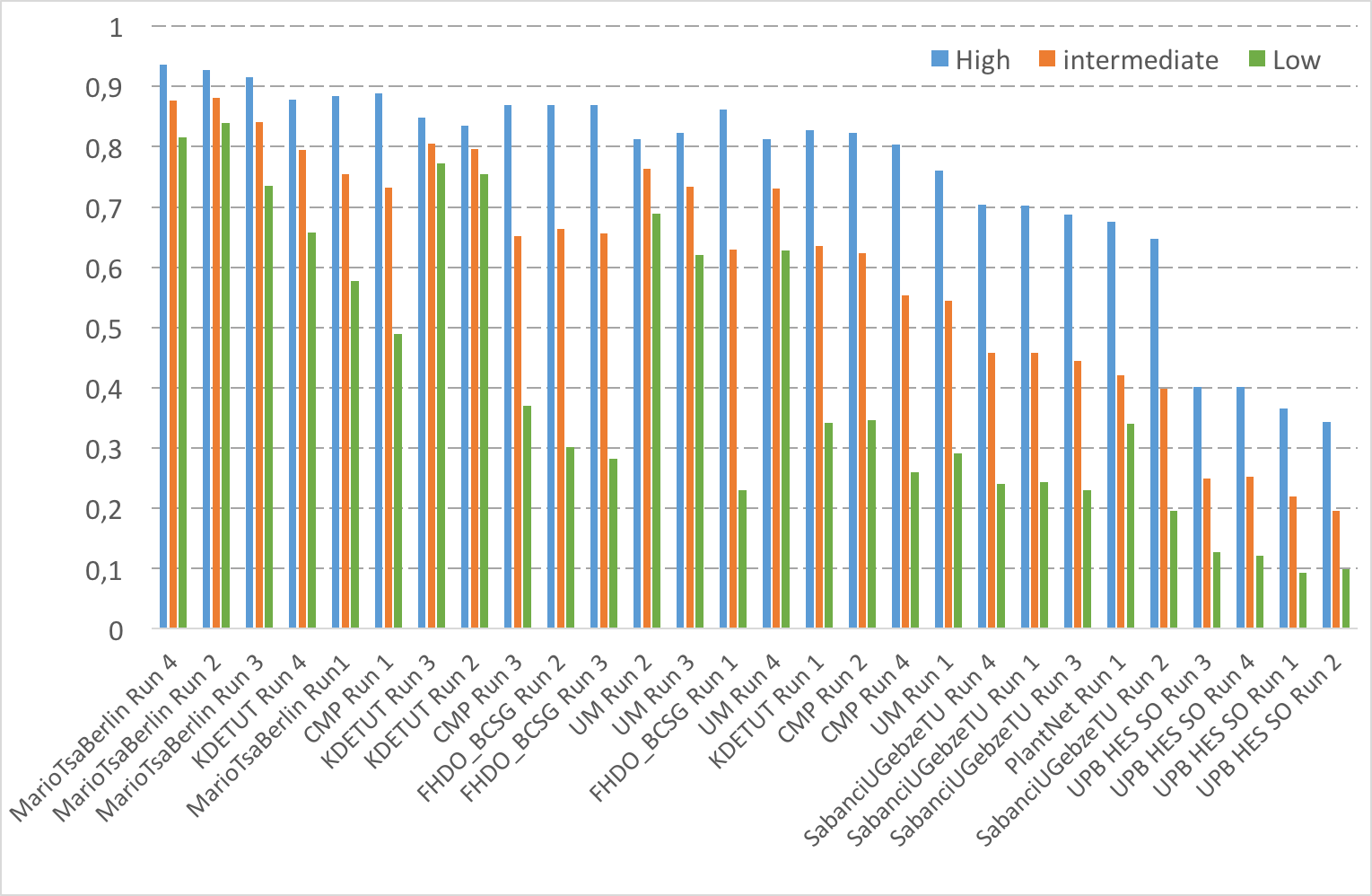}
\caption{MRR scores detailed by 3 categories of species: with an high, intermediate and low number of training images}
\label{fig:PlantCLEF2017ScoresByClassSize}
\end{figure}

\textbf{Performances by organs:} the main idea here is to evaluate which kind of organ and association of organs provide the best performances. The figure \ref{fig:PlantCLEF2017ScoresOrgans} gives the min, max and average MRR scores for all runs detailed by the 10 most representative organ combinations (with at least 100 observations in the test dataset). Surprisingly, the graph reveals that the majority of organ combinations share more or less the same MRR scores around 0.7 on average, highlighting how much the systems based on CNNs tend to be robust to any combination of pictures. However, as we yet noticed the previous years in LifeCLEF, the majority o f the systems performed clearly better when a test observation contains one or several picture of flowers exclusively. Using at least a picture of flower in a test observation with other types of organs guaranties in a sense good identification performances if we look at the three next organ combinations (flower-fruit-leaf, flower-leaf and flower-leaf-stem). On the opposite side, the systems have more difficulties when a test observation contains pictures of leaves without any flowers. It is getting worse when an observation combines only pictures of leaves and stems. This could be explained by the fact that stems are visually very different from leaves and both these two kinds of pictures produce dissimilar and non complementary sets of species on the outputs of the CNNs. We can notice as a complementary remark, that generally combining several pictures from different types of organs causes wider ranges of min and max scores, highlighting how much can be sensitive the combination of organs with an inappropriate rule.
\\
\\
\textbf{Biodiversity-friendly evaluation metric:} Like in \cite{Joly2014Biodiv} the main idea here is to evaluate how automated identification system deals with the long tail problem, \textit{i.e} how much an automated system performs along the long-tailed distribution, where basically very few species are well populated in terms of observations while the vast majority of species contain few images. We therefore split the species into 3 categories according to the number of observations populating these species in the datasets. The figure \ref{fig:PlantCLEF2017ScoresByClassSize} gives the detailed MRR scores of three categories of species: with a low, intermediate and high number of (trusted and noisy) training images, respectively between 4 and 161 images, between 162 and 195 images, and between 196 and 1583 images (the three categories are balanced in terms of total number of training images). First we can notice that, as we can expect, for the majority of the systems, the performances are clearly lower on the "intermediate" species than the "high" species, and even more on the "low" species category. For instance, the FHDO BCSG Run3 is very affected by the long tail distribution problem with a difference of MRR scores about 0.5 between the "high" and the "low" categories. However, on the opposite side, some runs like Mario TSA Berlin runs 2\&4, KDE TUT runs 2\&3, or to a lesser extent UM run 2, are definitely "biodiversity-friendly" since they are quite few affected by the long tail distribution and are able to maintain more or less equivalent MRR scores for the three species categories. We can specifically highlight the Run 2 from KDE TUT which, while it is using "only" three ResNet-50 models learned from scratch, is able to guaranty a MRR score around 0.79$\pm$0.4 almost independently from the number of training images by species. Moreover, we can notice that all these remarkable runs produced by KDE TUT, Mario TSA Berlin and UM teams, share the fact they used all the entire noisy datasets without any filtering process. All the attempts of filtering the noisy dataset seem to degrade the performances on the "intermediate" and "low" categories, like for instance for the Mario TSA Berlin Run 2 and 3 (resp. without an with filtering). 




\section{Conclusion}
This paper presented the overview and the results of the LifeCLEF 2017 plant identification challenge following the six previous ones conducted within CLEF evaluation forum. This year the task was performed on the biggest plant images dataset ever published in the literature. This dataset was composed of two distinct sources: a trusted set built from the online Encyclopedia of Life and a noisy dataset illustrating the same 10K species with more than 1M images crawled from the web without any filtering. The main conclusion of our evaluation was that convolutional neural networks (CNN) appear to be amazingly effective in the presence of noise in the training set. All networks trained solely on the noisy dataset did outperform the same models trained on the trusted data. Even at a constant number of training iterations (\textit{i.e.} at a constant number of images passed to the network), it was more profitable to use the noisy training data. This means that diversity in the training data is a key factor to improve the generalization ability of deep learning. The noise itself seems to act as a regularization of the model. Beyond technical aspects, this conclusion is of high importance in botany and biodiversity informatics in general. Data quality and data validation issues are of crucial importance in these fields and our conclusion is somehow disruptive.

\bibliographystyle{splncs03}

\end{document}